# APPLYING ARTIFICIAL NEURAL NETWORKS TO PREDICT NOMINAL VEHICLE PERFORMANCE

**Adam J. Last**
Applied Research Laboratory
P.O. Box 30
State College, PA 16804
e-mail: ajl5263@arl.psu.edu

**Timothy F. Miller**
Applied Research Laboratory
P.O. Box 30
State College, PA 16804
e-mail: nfn@arl.psu.edu

## ABSTRACT

This paper investigates the use of artificial neural networks (ANNs) to replace traditional algorithms and manual review for identifying anomalies in vehicle run data. The specific data used for this study is from undersea vehicle qualification tests. Such data is highly non-linear, therefore traditional algorithms are not adequate and manual review is time consuming. By using ANNs to predict nominal vehicle performance based solely on information available pre-run, vehicle deviation from expected performance can be automatically identified in the post-run data. Such capability is only now becoming available due to the rapid increase in understanding of ANN framework and available computing power in the past decade. The ANN trained for the purpose of this investigation is relatively simple, to keep the computing requirements within the parameters of a modern desktop PC. This ANN showed potential in predicting vehicle performance, particularly during transient events within the run data. However, there were also several performance cases, such as steady state operation and cases which did not have sufficient training data, where the ANN showed deficiencies. It is expected that as computational power becomes more readily available, ANN understanding matures, and more training data is acquired from real world tests, the performance predictions of the ANN will surpass traditional algorithms and manual human review.

## PROBLEM

Undersea vehicles, like any high performance systems, require many qualification runs prior to and throughout their service lives to ensure proper functionality. Results from these qualification runs can include hundreds of channels of data, including speeds, pressures, temperatures, flow rates, vibrations, and many others. Further, each qualification run usually has a unique mission profile, which is the progression of commanded speeds throughout the run. This makes direct comparison between qualification runs difficult.

Therefore, identifying anomalous behavior from the post-run data requires many highly trained and experienced engineers manually reviewing each data channel. Even then, due to human error, subtle anomalies can be overlooked or, conversely, normal behavior can be flagged as anomalous. To counter this, creating an algorithm or method to predict results based solely on the mission profile is highly desirable. The current best method is to average the measured values for a given data channel at a given speed from all previous runs, and assume that is the “normal” operating condition for that speed. However, the response of each data channel is highly non-linear, as not only the current speed command but also the cumulative effect of all prior operation determines current performance. As a result, traditional algorithms become unwieldy and unreliable for this application.

## INTRODUCTION

Overcoming the breakdown of traditional algorithms on large non-linear data sets has been a focus for many industries in recent years. Data center optimization [1], handwriting recognition [2], language translation [3], and image classification [4] are examples of problems that traditional algorithms cannot effectively solve. Decades of exponential growth in available computing power, along with recent developments in mathematical methods, has enabled a new approach to handling large data sets- artificial neural networks (ANNs). ANNs have been deployed in the last few years to enable everything from real time language translation [5] to instant handwriting recognition [6] to financial market analysis [7].

This paper outlines the approach to using the existing undersea vehicle qualification run data to train an ANN to predict future hypothetical run results, with the only input being the mission profile. Then, when the hypothetical qualification run is actually performed in the real world, the results can be fed back to the computer. The ANN can then identify where differences exist between its predicted result and the actual results to automatically flag anomalous behavior.

## TECHNICAL BACKGROUND

The first work on artificial neural networks began in the mid-20th century with the concept of perceptrons. The perceptron is essentially a logic gate, where multiple inputs are taken in, individually weighted, and summed. If the sum of the weighted inputs are greater than a threshold value, then the output of the perceptron is “1”, otherwise the output is “0”. This is shown graphically in Figure 1 (graphic adapted from Nielson [8]).

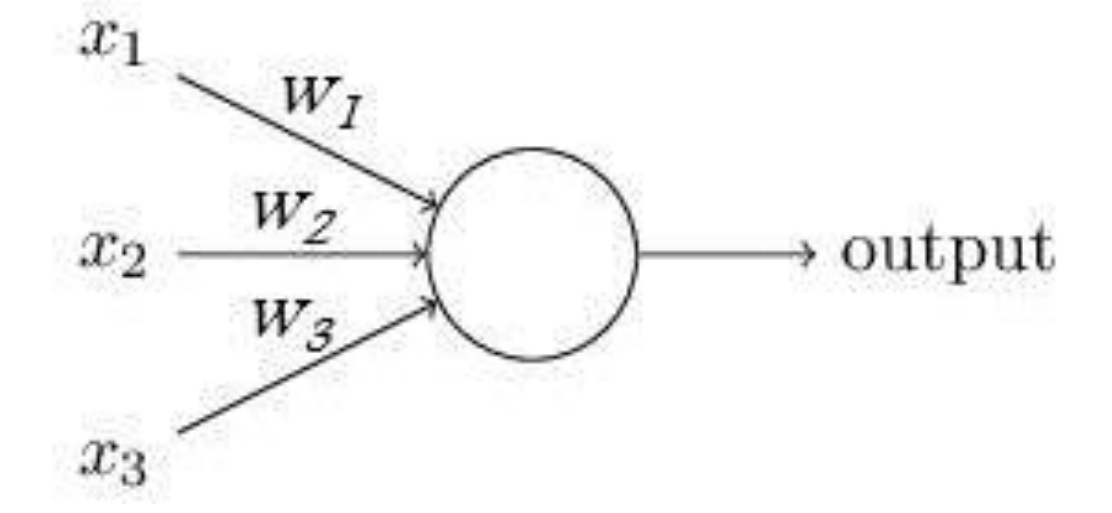

$$output = \begin{cases} 0\ if \sum w_j x_j \le threshold \\ 1\ if \sum w_j x_j > threshold \end{cases}$$

*Figure 1. Perceptron logic*

The perceptron was later refined to allow for non-binary output (i.e. output other than just 1 and 0). This is accomplished by replacing the step function logic with a sigmoid function. For this reason, the new model for the perceptron was named the “sigmoid neuron”. The “neuron” portion of the name derives from the sigmoid neuron’s similarity in operation to a biological neuron, which on a very basic level activates an output given a sufficient input [8].

Sigmoid neurons become useful when linked together to perform complex logic, and were renamed “hidden neurons” in the scope of neural networks. Figure 2 (graphic adapted from Nielson [8]) shows an artificial neural network containing 3 inputs, 4 hidden neurons, and one output.

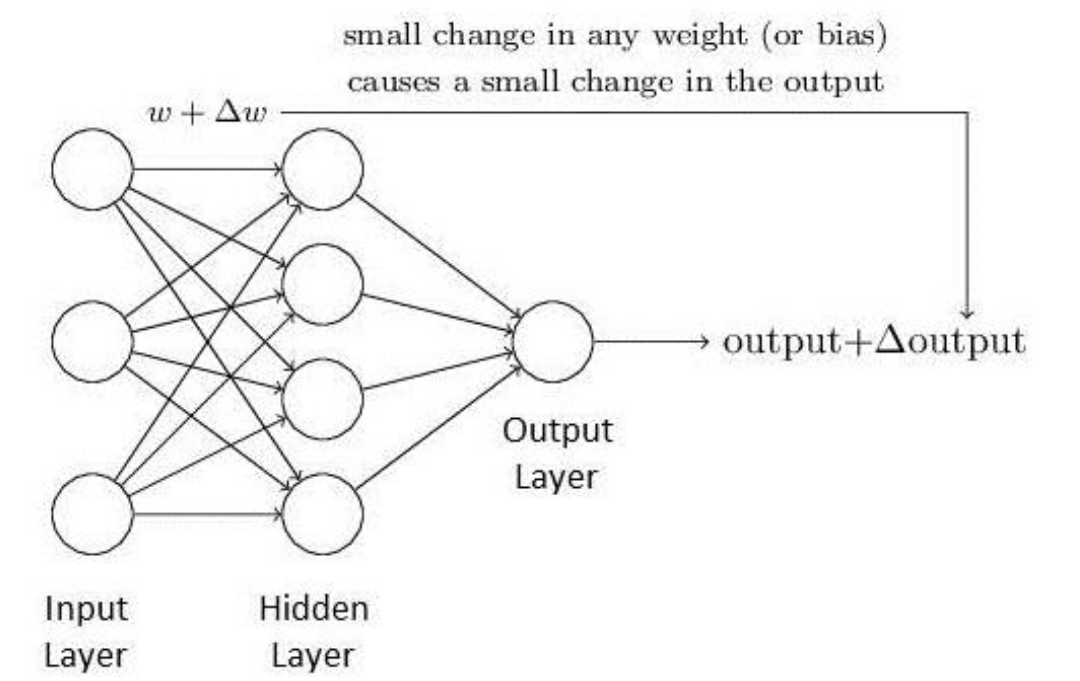

*Figure 2. Simple ANN*

To orient Figure 2 to the current problem at hand, the input layer would be known values (such as the commanded speed). The output layer would be whatever data channel you are trying to simulate, the most intuitive being actual vehicle speed. There will obviously need to be far more than 4 hidden neurons for any meaningful results.

For a neural network to be useful, the weighting factors for each input to each hidden layer sigmoid neuron must be defined. To do so, a training process must be undertaken. The training process requires a large set of already known input/output pairings. In this case, this known set is the available previous vehicle qualification runs, which have known inputs (such as commanded speed) and measured output (such as actual speed). Initially in the training process, a random set of weights is defined. The output is then calculated and compared to the actual output. This is the error in the neural network for the current “evolution”. The weights are then adjusted (shown as $\Delta w$ in Figure 2) and the ANN resolved. This process is repeated until the error is below an acceptable value or a preset number of evolutions has been achieved [9].

There are several methods for determining the amounts by which to adjust each individual weight. One of the easiest to implement and most computationally efficient is the optimization method of stochastic gradient decent. Gradient descent is a well-known mathematical model for finding minima and maxima. Stochastic methods are employed to introduce an element of randomness into the optimization routine. This allows for the optimization to escape local minima and maxima during the solution and continue improving on the solution through many evolutions [10].

While stochastic gradient decent is efficient, it is still very computationally intensive when solving the weights for many (hundreds or more) hidden layer sigmoid neurons. That is why the recent sustained exponential growth in available computing power has been critical to making ANNs practical. Once the ANN weights are solved, it takes very little computational resources to run new inputs through the ANN to find the output.

There have been several advances in mathematical methods in recent years that have made ANNs much more powerful and extensible to more applications. However, for the purpose of this paper, using the methods already outlined is sufficient.

## ANN SETUP

A simple ANN was set up that was capable of being run on a single desktop computer with 4 GB RAM. This ANN was built using the Netlab MATLAB toolbox developed by Nabney [11]. The training data was taken from 20 qualification runs of an undersea vehicle. This gave 64,779 unique "time steps". Each "time step" represents a discrete amount of runtime. Time and speeds have been non-dimensionalized.

For each "time step", input values must be provided to the ANN. For the engine of this vehicle, like many other vehicles, the only true control input is commanded speed. All other engine parameters are adjusted internally for the goal of matching that commanded speed as closely as possible. Though not strictly a "control" parameter, the other engine input is the amount of energy reserves remaining (and, by extension, the amount of reserves that has already been used). For various systems, this could take the form of percentage of combustible fuel remaining in a tank, remaining battery power, etc. For many systems, such as the one under study, this energy reserves input also functions as a "state of health" parameter (i.e. for electric systems, motors may not perform optimally once the batteries drop below a threshold voltage).

From these general "inputs", the specific input data that was deemed pertinent to be provided to the ANN is as follows…

1. Speed Command Information…
    a. Current Commanded Speed
    b. Time at Current Commanded Speed
    c. Previous Commanded Speed
2. % of Energy Reserves Depleted (termed "Utilization" for the purposes of this paper)

Also provided for training the ANN was the actual measured vehicle speed for each data point. The "success" of the ANN training would be determined by comparing the actual vehicle measured speed to the ANNs predicted vehicle speed.

The ANN was initialized to have 750 hidden sigmoid neurons. The training was set to terminate after 750,000 evolutions. This is about 48 hours of computation time on a PC with an Intel Core i5-2400 CPU running at 3.1 GHz.

To assist with visualization of the training process of this specific ANN, the notation from Figure 2 is used in Figure 3 below, depicting the ANN setup. The routine of Figure 3 would be run once for each time step each evolution (64,779 times every evolution for this data set).

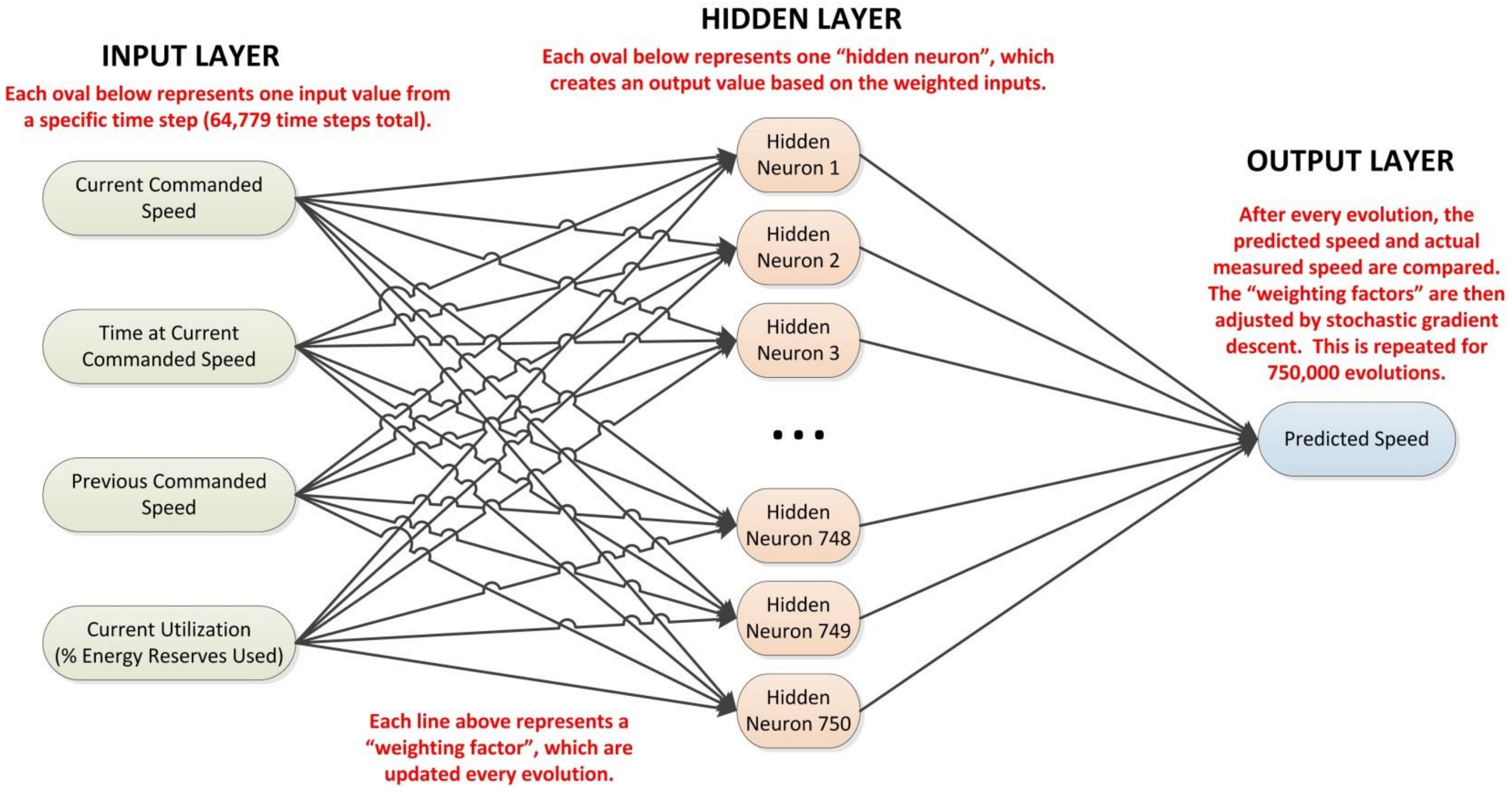

*Figure 3. ANN setup and training visualization*

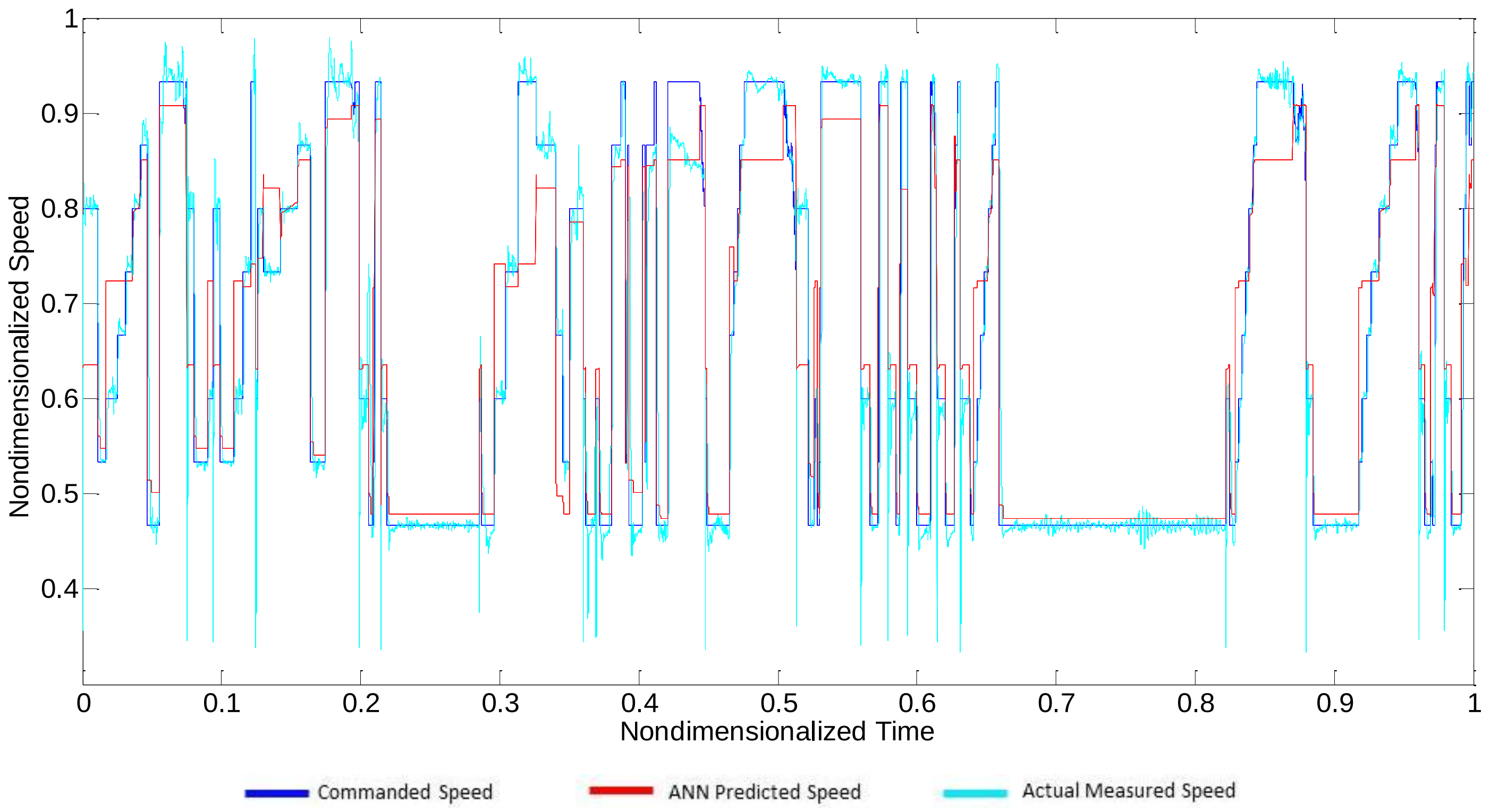

*Figure 4. Full ANN results*

## ANN RESULTS

Figure 4 above shows the commanded speed (dark blue line) and the ANN's prediction for actual speed (red line). Also given is the true actual speed as measured in the qualification tests (light blue line).Quick inspection of Figure 4 shows that the ANN has largely succeeded in matching the general speed response. However, there are several spans of time where the ANN fails to predict any more accurately than simply guessing that the actual speed is equal to the commanded speed. The most obvious example is during long periods of constant speed, known as steady state. Conversely, there are several spans where the ANN prediction is much more accurate than any traditional algorithm could be, such as during speed changes (transients)

It is critical to attempt to understand the reasons why the ANN is very accurate or not accurate in various situations. If these reasons can be well understood, then adjustments can be made to the ANN training routine to address deficiencies and promote strengths. Examples would be providing additional input information, working to obtain test data for situations where the ANN shows deficiencies, changing the number of hidden neurons, or adjusting the evolutions to simulation completion. To that end, the remainder of this section of the paper examines three specific performance cases where the ANN shows interesting prediction behavior. The upcoming Figures 5, 6, and 7 all show magnified subsets of data from Figure 4, to allow for in-depth analysis.

### *Consistently low prediction of high speed command performance*

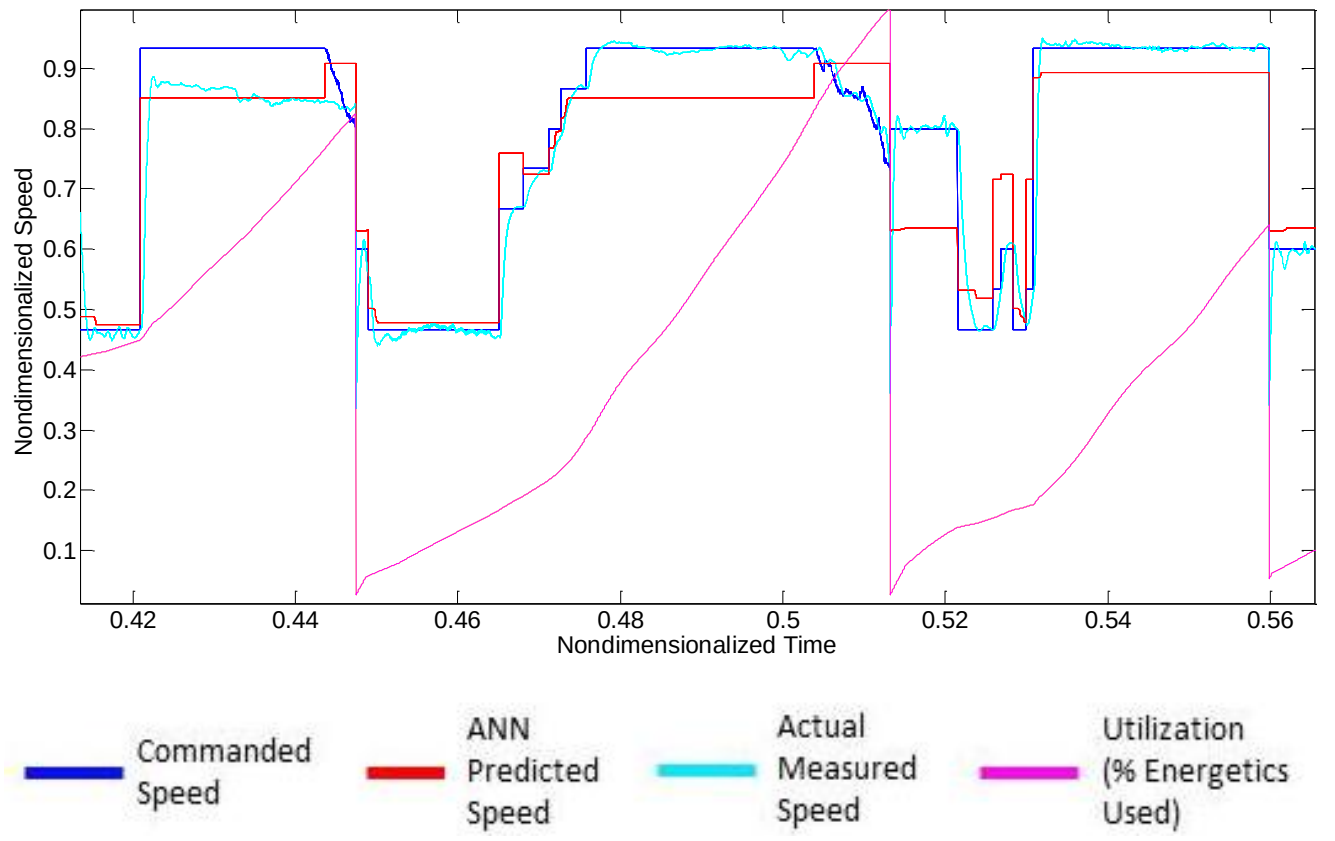

*Figure 5. ANN results, low predictions of high speed operation*

Figure 5 shows a magnified portion of the ANN data of Figure 4, from 0.42 to 0.56 of the nondimensional time. Added to the plot, that was not present in Figure 4, is the non-dimensionalized utilization (% of energetics used), shown in magenta. Note that there are three separate "runs" shown, as the utilization resetting to zero is indicative of a new run. In this timeframe there are three instances of the vehicle being commanded to its highest speed. In all three instances, the ANN predicts the actual speed will only achieve about 80% of the full speed value. However, this is only true for one of the three occurrences; in the other two the vehicle does achieve high speed operation. It is probable that the ANN is chronically under predicting the actual high speed operation as a “safe” prediction. This is because the speed sometimes undershoots, but almost never overshoots, a high speed command. The likelihood that the vehicle will undershoot the high speed command is increased at high vehicle utilization (near the end of an individual run- because effective control is more difficult), as it was in the one instance of undershooting in Figure 5. The ANN shows slight sensitivity to this as the high speed command in Figure 5 furthest to the right occurs at the lowest utilization; the ANN displays the highest predicted speed. However, the prediction is still too low, and the ANN clearly does not fully incorporate the link between utilization and the potential for undershooting. More run data which contains more high speed command operation at various utilizations would likely remedy this issue.

### *Accurate prediction of transient performance*

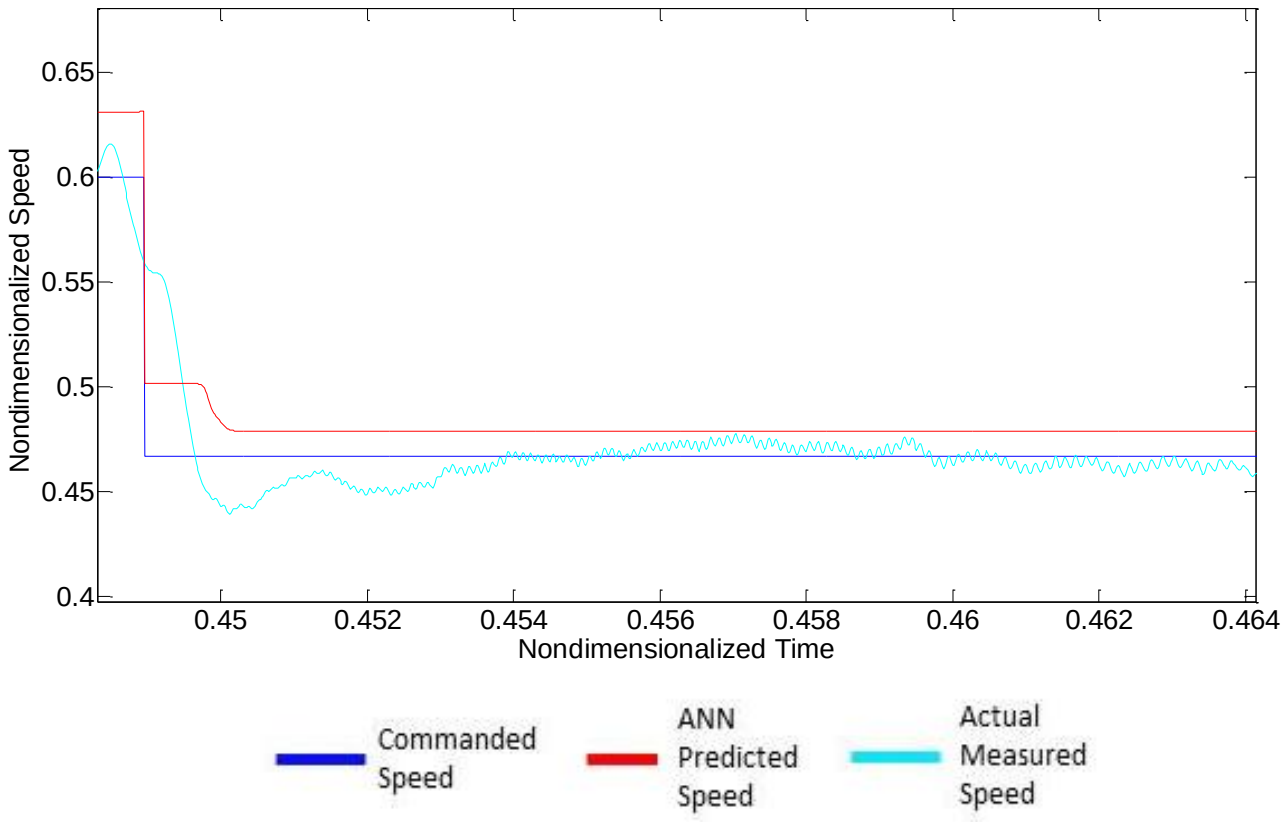

*Figure 6. ANN results, transient performance*

Figure 6 shows one of the speed command changes in the data. During the transient, the ANN predicted speed showed a sharp drop then gradual step down to the final steady state speed, consistent with actual vehicle operation. This is a feature that a traditional algorithm would have trouble matching, particularly once data other than speed (temperatures, pressures, etc.), which may take a much longer time than speed to reach steady state, are considered.

### *Data overfitting during “staircase” speed increase*

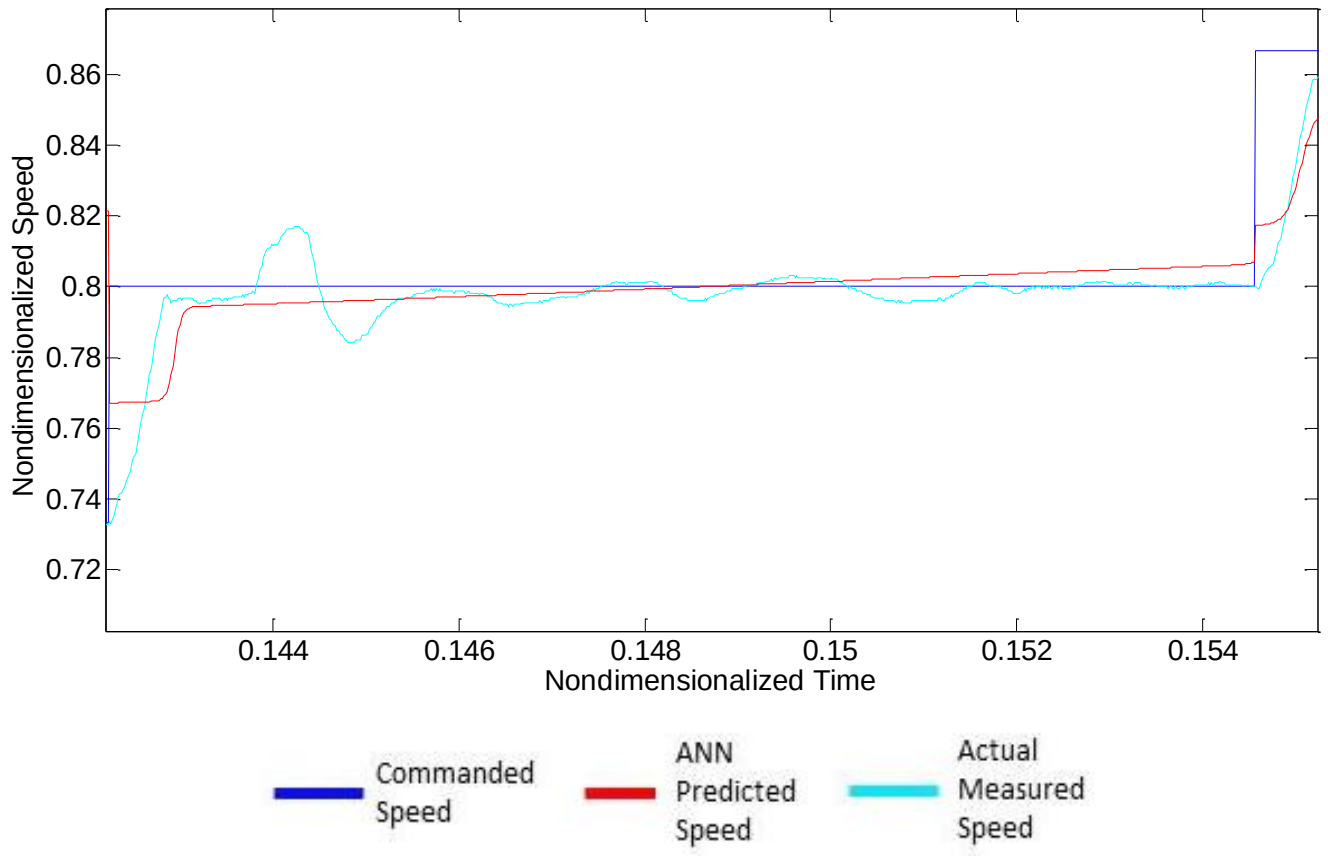

*Figure 7. ANN results, overfit solution*

Figure 7 shows a brief speed command of 0.8 non-dimensionalized speed. The ANN predicted speed appears to match the actual speed of the vehicle closely over this time. However, the predicted speed is slowly increasing the entire time, as if “anticipating” the next speed command. The reason is that this speed change is part of a “staircase” pattern of speed changes, where the speed goes from the minimum to the maximum over a series of small changes in speed in rapid succession. These “staircase” patterns are very common in the training run data. Therefore, the ANN learned that when the speed changes from a moderate speed to just a slightly higher speed, in all likelihood it was in the middle of a “staircase” speed profile and should anticipate another small speed step up. However, it is obviously not always true that these small speed steps must be part of a “staircase” sequence. Therefore, the “anticipation” of the ANN for another immediate speed change is unwarranted and indicative of overfitting of the data [12]. This would be remedied by additional run data which showed small speed steps which are not part of a “staircase” pattern.

## CONCLUSIONS

The results show that even simple ANNs show promise of outperforming traditional prediction methods for times in the data which are highly non-linear (such as speed transients). However, the simple ANN does not have adequate performance for all times, most notably during steady state operation. It is highly probably that the ANN would meet or exceed the performance of traditional algorithms if a more complex ANN were solved on a more powerful computer.

A more fundamental limitation of the ANN performance may be the potential for overfitting of data. This is a result of not having a sufficiently large set of training data. Specifically, if the same sequence of speed steps is present in most data sets, the ANN will be highly accurate in predicting results for that exact sequence, but not when the sequence is altered slightly. Therefore, it is possible that even with an advanced ANN solved on a powerful computer, the size of the training data set could prove a limitation to ANN effectiveness. The mitigation for this problem is the inclusion of additional data sets in the training data, especially those with a diverse mission profiles.

## FUTURE WORK

The size of the training data set continues to increase, as qualification runs of the vehicle are ongoing. Additionally, computational power continues to become more and more readily available. Therefore, it is prudent to develop an advanced ANN framework now and run it with the existing data set with the understanding that the results may not yet outperform traditional algorithms or manual data analysis. However, with that framework in place, retraining the ANN with new data becomes a trivial exercise. As more data and computational power becomes available it can be expected that at some point the ANN will be provide the best predictive capability. At that point additional data channels (temperatures, pressures, etc.) can all be predicted using the same ANN framework.

Once ANNs have been proven to reliably identify anomalies from post run data, it will be possible to extend their use into identifying anomalies during a run itself. For instance, if a vehicle behaves off-prediction by some margin, the ANN can alert the vehicle control which can then evaluate whether to issue an early termination of the run. Even further in the future, ANNs should be capable of assuming control for adjusting vehicle run parameters (flow rates, pressures, etc.) in real time to optimize vehicle performance.

## CLOSING REMARKS

The conclusions of this paper (requiring more computational power and better mathematical techniques, but more importantly additional training data) are consistent with the needs of nearly all ANN projects currently in development.

On the computational side, chipmakers are rushing to provide hardware designed specifically for ANN training and deployment [13]. This hardware leverages architecture traditionally used for intensive graphics processing. Brand new architecture is under development which will be even further optimized for machine learning and neural networks [14].

On the training data side, companies have been using devices owned and used by individuals to collect data to improve their algorithms. The most relevant of these to this paper might be Tesla Motors, which receives data from every one of its cars on the road to improve its “autopilot” (autonomous driving) feature [15]. Interestingly, this type of training data collection is a driving force in the recent public debate about the privacy of personal electronic devices. For instance, Google’s efforts to track cell phone activity stems from their desire to expand the training sets for the algorithms that control search result prioritization, traffic congestion prediction, language translation, and more [16].

## NOMENCLATURE

$x$ = ANN training data input
$w$ = ANN sigmoid neuron weight

## ACKNOWLEDGMENTS

The data for this paper was provided by ARL PSU from experiments performed under NAVSEA contract. We thank the entirety of ARL’s Energy Science and Power Systems division for the support and resources. The Netlab MATLAB toolbox which was utilized to perform the ANN training was developed at Aston University by Ian Nabney and Christopher Bishop.

## REFERENCES

[1] Gao, Jim. Machine Learning Applications for Data Center Optimization. Rep. Mountain View: Google, 2013. Web. <https://docs.google.com/a/google.com/viewer?url=www.google.com/about/datacenters/efficiency/internal/assets/machine-learning-applicationsfor-datacenter-optimization-finalv2.pdf>.

[2] Le Cun, Y. "Handwritten Character Recognition Using Neural Network Architectures." 4th USPS Advanced Technology Conference (1990): n. pag. AT&T Bell Laboratories. Web. 20 Oct. 2015. <http://yann.lecun.com/exdb/publis/pdf/matan-90.pdf>.

[3] Auli, Michael. "Joint Language and Translation Modeling with Recurrent Neural Networks." Microsoft Research (n.d.): n. pag. Microsoft. Web. 20 Oct. 2015. <http://research-srv.microsoft.com/en-us/um/people/gzweig/Pubs/EMNLP2013RNNMT.pdf>.

[4] Szegedy, Christian. "Building a Deeper Understanding of Images." Research Blog. Google, n.d. Web. 20 Oct. 2015. <http://googleresearch.blogspot.com/2014/09/building-deeper-understanding-of-images.html#uds-search-results>.

[5] Bergen, Mark. "Google's Neural Nets Can Now Translate Text Instantly, Practically Everywhere." Recode. Recode, 29 July 2015. Web. 20 Oct. 2015. <http://recode.net/2015/07/29/googles-neural-nets-can-now-translate-text-instantly-practically-everywhere/>.

[6] Martendale, Jon. "Now Our Computers Can Forge Our Letters and Checks." Digital Trends. Digital Trends, 22 July 2015. Web. 20 Oct. 2015. <http://www.digitaltrends.com/computing/researcher-computer-handwriting/>.

[7] "Neural Networks - Applications." Neural Networks - Applications. N.p., n.d. Web. 20 Oct. 2015. <http://cs.stanford.edu/people/eroberts/courses/soco/projects/neural-networks/Applications/stocks.html>.

[8] Nielson, Michael A. Neural Networks and Deep Learning. N.p.: Determination, 2015. Determination Press. Web. 20 Oct. 2015. <http://neuralnetworksanddeeplearning.com/chap1.html>.

[9] Bishop, Christopher M. Neural Networks for Pattern Recognition. Oxford: Clarendon, 1995. Print.

[10] Bottou, Leon. "Stochastic Gradient Descent Tricks." Microsoft Esearch (n.d.): n. pag. Web. 20 Oct. 2015. <http://research.microsoft.com/pubs/192769/tricks-2012.pdf>.

[11] Nabney, Ian T. NETLAB: Algorithms for Pattern Recognition. London: Springer, 2003. Print.

[12] "Documentation." Improve Neural Network Generalization and Avoid Overfitting. Mathworks, n.d. Web. 20 Oct. 2015. <http://www.mathworks.com/help/nnet/ug/improve-neural-network-generalization-and-avoid-overfitting.html>.

[13] Ferrucci, David. "AI MAGAZINE." Building Watson: An Overview of the DeepQA Project. AAAI, Aug. 2007. Web. 03 Feb. 2016. <http://www.aaai.org/ojs/index.php/aimagazine/article/view/2303/2165>.

[14] Colaner, Seth. "Movidius, Google Team Up To Leverage Deep Learning On Mobile Devices." Tom's Hardware. Tom's Hardware, 27 Jan. 2016. Web. 3 Feb. 2016. <http://www.tomshardware.com/news/movidius-google-deep-learning-mobile-devices,31078.html>.

[15] Fehrenbacher, Katie. "How Tesla Is Ushering in the Age of the Learning Car." How Tesla Is Ushering in the Age of the Learning Car Comments. Fortune, 16 Oct. 2015. Web. 03 Feb. 2016. <http://fortune.com/2015/10/16/how-tesla-autopilot-learns/>.

[16] Lafrance, Adrienne. "The Convenience-Surveillance Tradeoff." The Atlantic. Atlantic Media Company, 14 Jan. 2016. Web. 03 Feb. 2016. <http://www.theatlantic.com/technology/archive/2016/01/the-convenience-surveillance-tradeoff/423891/>.

## ANNEX A

## INPUT FILE TO NETLAB TOOLBOX (MATLAB CODE)

```
% Running this script requires installation of Netlab 3.3 MATLAB toolbox
% Netlab toolbox available at
http://www.aston.ac.uk/eas/research/groups/ncrg/resources/netlab/downloads/
% Script derived from standard Netlab input file, demmlp1.m, by Ian T Nabney

clear all;
clc

% Input arrays must be initialized to variable 'x' as follows...
%    First Column = Current Commanded Speed
%    Second Column = Previous Commanded Speed (0 if run just started)
%    Third Column = Time Since Last Speed Command Change
%    Fourth Column = Current Utilization

% Set up network parameters.
nin = 4; % Number of inputs. (see input array comment above)
nhidden = 750; % Number of hidden sigmoid neurons.
nout = 1; % Number of outputs.
alpha = .1; % Coefficient of weight-decay prior.

% Create and initialize network weight vector.
net = mlp(nin, nhidden, nout, 'linear', alpha);

% Set up vector of options for the optimiser.
options = zeros(1,18);
options(1) = 1; % This provides display of error values.
options(14) = 75000; % Number of training cycles.

% Run Netlab to train neural network
[net, options] = netopt(net, options, x, t, 'scg');

% Use neural network to generate predicted speeds
y = mlpfwd(net, x);
```

## ANNEX B

## VALIDATION SET

Several months after the ANN described in this paper had been trained, new field test data became available. This new data was used as a "validation set", in order to test the efficacy of the ANN without re-training. The results are shown in Figure B-1 below.

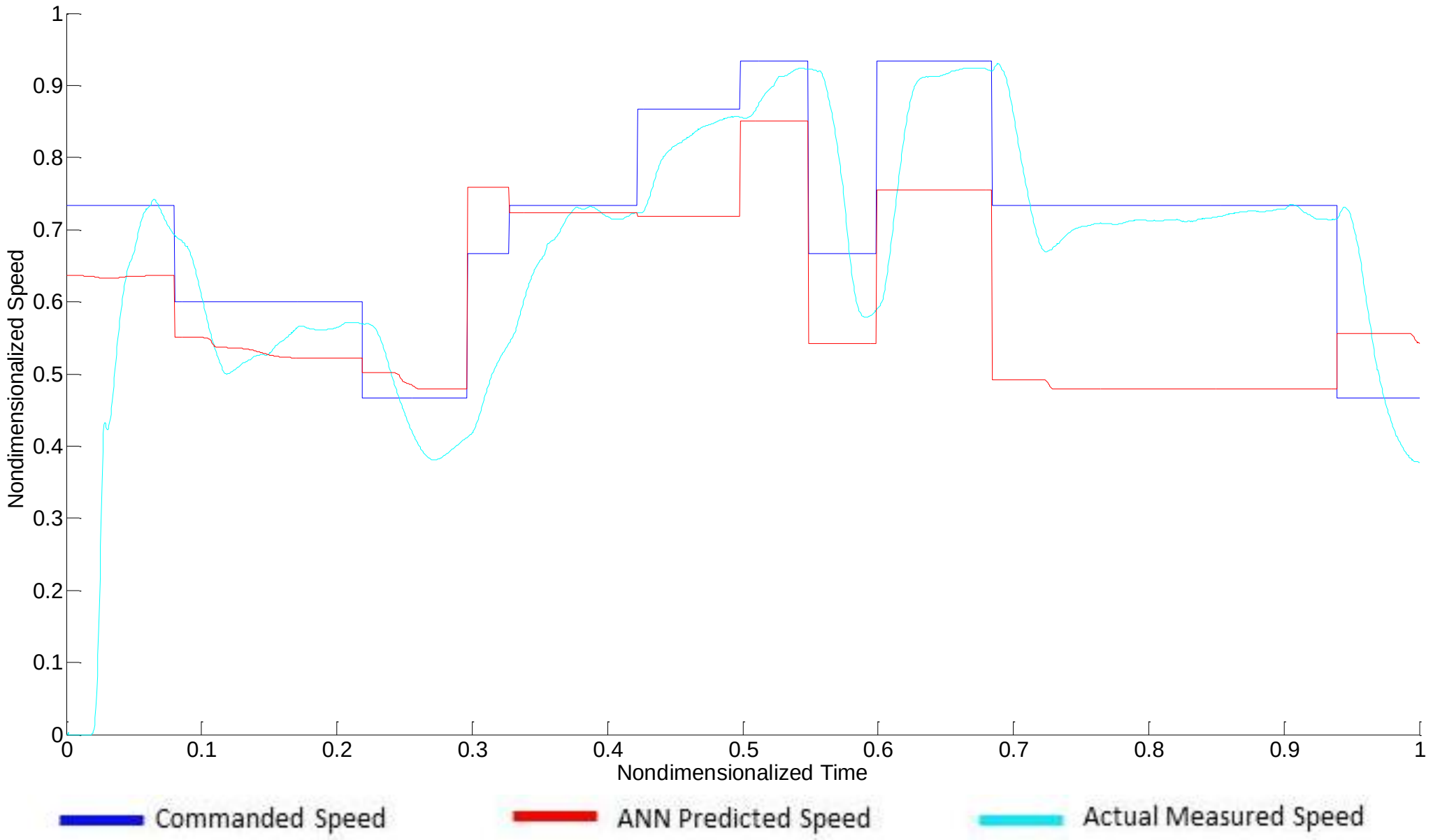

*Figure B-1. Validation Set Results*

A few observations from the ANN results are given below.

- *Non-dimensionalized time = 0*. The ANN is unable to capture the speed change at "startup". This is likely because there are relatively few "startups" in the training data (there is only 1 startup per field test, with 20 field tests used in the training data).
- *Non-dimensionalized time = 0.25*. Through this speed step, the ANN predicts a near-constant speed that is the average of the speed transient. This is likely due to an insufficiently advance algorithm, as the ANN "settled" into minimizing error by picking a mid-point speed, rather than accurately modeling the transient.
- *Non-dimensionalized time = 0.3*. The ANN anticipated a speed "overshoot" during this transient, which did not occur. Analysis of the training data shows that, on several occasions, an "overshoot" did happen at this speed. Therefore, this ANN error is the results of an insufficient number of training sets that do not have an overshoot during a similar speed step.
- *Non-dimensionalized time = 0.55*. The ANN correctly predicts a speed "undershoot" during this transient.
- *Non-dimensionalized time = 0.6 through end of run*. The ANN predicts that the speed will be much less than the actual speed through the end of the run. This is likely due to previous runs in the training set that have failed to maintain speed late in run. Therefore, more training data is necessary for the ANN to be able to distinguish the characteristics of runs that will and will not be able to maintain speed.

From the results, we can calculate the accumulated error using two speed prediction methods. The first method is simply presuming the actual speed is equal to the setpoint (this will be referred to as the "Non-ANN Method"). It should be noted that this method could be improved by simple algorithms, such as assuming a time constant for speed changes, but that is not the focus of this analysis. The second method is the prediction of the ANN. The error of each of these methods can be calculated as the absolute value of prediction minus the actual speed. A graph of accumulated error is shown in Figure B-2 below.

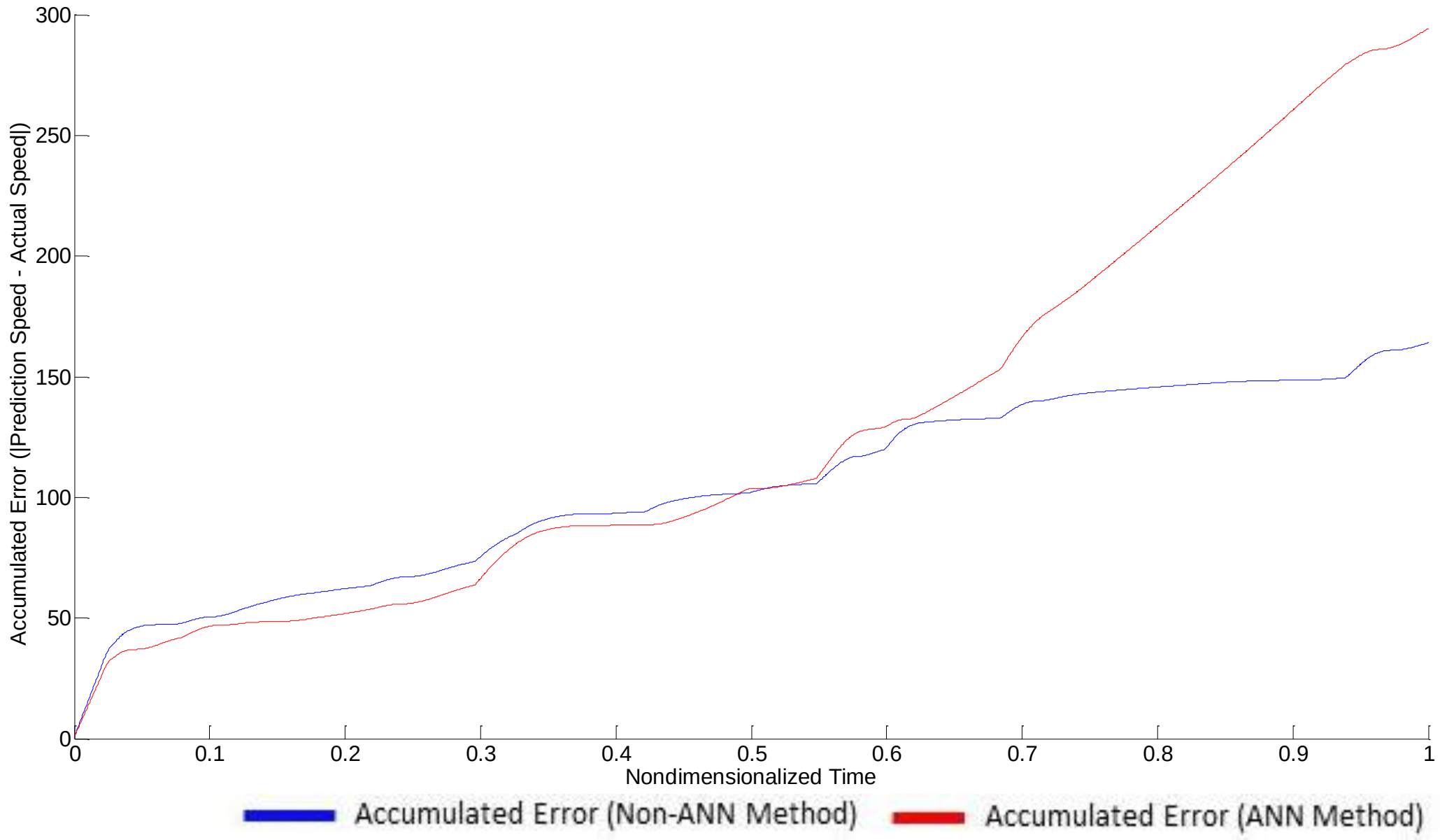

*Figure B-2. Accumulated Error from Non-ANN and ANN Speed Prediction Methods on Validation Set*

Figure B-2 shows that the Non-ANN and ANN methods for speed prediction have nearly the same error accumulation rate until late in the data set (second to last speed change). This is an encouraging sign that with additional training-set data and more advanced ANN methods, neural networks will be practical tools for predicting vehicle performance in the near future. In summary, the results of this validation set reinforce the assertions made in the conclusion of this paper that ANN performance can be improved with more advanced methods and more computing power, but more training data will be needed as well.